\documentclass[french,a4paper]{rnti}
\usepackage{indentfirst}
\usepackage{graphics}
\usepackage{subfigure,graphicx}
\usepackage{fancyhdr}
\usepackage{calc}
\usepackage{multirow}
\usepackage{wrapfig}

\usepackage{lscape}

\usepackage{amssymb,amsmath}
\usepackage{algorithm,algorithmic}
\usepackage{amssymb, amsthm, amsmath, amsfonts}

\newcommand{\conj}{\mathrm{Conj}}
\newcommand{\dis}{\mathrm{Dis}}
\newcommand{\pl}{\mathrm{pl}}
\newcommand{\bel}{\mathrm{bel}}
\DeclareMathOperator{\ocap}{\displaystyle{\small{\textcircled{{\scriptsize $\cap$}}}}}
\DeclareMathOperator{\owedge}{\displaystyle{\small{\textcircled{{\scriptsize $\wedge$}}}}}

\titrecourt{Intégration d'une mesure d'indépendance pour la fusion d'informations}

\nomcourt{Kharoune et Martin}

\titre{Intégration d'une mesure d'indépendance pour la fusion d'informations}

\auteur{Mouloud Kharoune\affil{1},
        Arnaud Martin\affil{1}}

\affiliation{
    \affil{1}UMR 6074 IRISA, Université de Rennes1 / IUT de Lannion, Rue Edouard Branly\\BP 3021, 22302 Lannion cedex\\
Mouloud.Kharoune@univ-rennes1.fr, Arnaud.Martin@univ-rennes1.fr }

\resume{%
La fusion d'informations fait intervenir plusieurs sources d'informations afin d'améliorer la décision en terme de certitude et de précision. Quelle que soit l'approche retenue pour réaliser la fusion d'informations, l'hypothèse d'indépendance est généralement une hypothèse forte et incontournable. Nous proposons dans cet article une approche permettant d'intégrer une mesure d'indépendance avant de réaliser la combinaison des informations dans le cadre de la théorie des fonctions de croyance.}

\summary{%
Many information sources are considered into data fusion in order to improve the decision in terms of uncertainty and imprecision. For each technique used for data fusion, the asumption on independance is usually made. We propose in this article an approach to take into acount an independance measure befor to make the combination of information in the context of the theory of belief functions.
}

\begin{document}
\section{Introduction}

Tel que repris par~\cite{Martin05c}, la fusion d'informations consiste à combiner des informations issues de plusieurs sources afin d'aider à la prise de décision. Les approches de fusion d'informations cherchent donc à tenir compte des redondances des informations issues des différentes sources. Les approches de fusion n'ont bien sûr d'intérêt que si les sources sont imparfaites et fournissent des informations peu sûres et précises qui se complètent. Ainsi il faut donc chercher à modéliser aux mieux les imperfections des sources et des données. Pour ce faire, différentes théories de l'incertain ont été sollicitées. Parmi elles, citons la théorie des probabilités, des sous-ensembles flous et des possibilités ou encore la théorie des fonctions de croyance. Quel que soit le cadre théorique retenu, lors de l'étape de combinaison de l'information, l'hypothèse d'indépendance des sources est généralement faite. Cette hypothèse est plus ou moins forte. Par exemple, l'indépendance statistique est généralement 
retenue pour appliquer plus aisément la combinaison bayésienne du cadre probabiliste. En effet, les estimations peuvent s'avérer très vite compliquées sans cette hypothèse. Dans le cas de la théorie des fonctions de croyance, il est question d'indépendance cognitive définie par~\cite{Shafer76}. Elle correspond à une absence de communication entre les sources sans que celles-ci soient pour autant indépendantes statistiquement. Malheureusement cette hypothèse d'indépendance est rarement vérifiée ou justifiée.

Dans ce travail nous nous intéressons plus particulièrement à la théorie des fonctions de croyance car elle offre un outil riche de modélisation et de gestion de l'information. En particulier, \cite{Chebbah12a,Chebbah13a} ont récemment proposé des approches pour mesurer l'indépendance des sources en distinguant de plus la dépendance positive et négative. 

Nous poursuivons ainsi cet article en présentant les principes de base de la théorie des fonctions de croyance et en particulier la notion d'indépendance et d'affaiblissement. Nous exposons ensuite l'approche proposée permettant de tenir compte d'une mesure d'indépendance avant la combinaison des fonctions de croyance. Nous illustrons ce principe à partir d'exemples générés.

\section{Théorie des fonctions de croyance} 
La théorie des fonctions de croyance issue des travaux de \cite{Dempster67a}, repris par \cite{Shafer76} est depuis quelques années employée dans des applications de fusion d'informations. Nous présentons ci-dessous les principes de cette théorie.

\subsection{Principes de base}

Considérons le cadre de discernement $\Omega=\{\omega_1, \omega_2,\ldots, \omega_n\}$ correspondant à l'ensemble de toutes les hypothèses possibles de décision d'un problème donné. Les éléments $\omega_i$ représentent ainsi toutes les hypothèses exclusives et exhaustives. 

L'ensemble $2^\Omega=\{A/A\subseteq\Omega\}=\{\emptyset, \omega_1, \omega_2,\ldots, \omega_n, \omega_1\cup\omega_2, \ldots, \Omega\}$, est composé de toutes les disjonctions de $\Omega$. L'espace puissance $2^\Omega$ comporte $2^{|\Omega|}=2^n$ éléments.

Une fonction de masse est une fonction de $2^\Omega$ vers l'intervalle $[0,1]$ qui affecte à chaque sous-ensemble de $2^\Omega$ une valeur de l'intervalle $[0,1]$ représentant sa masse de croyance élémentaire. Elle s'écrit~:
\begin{equation}
m^\Omega:2^\Omega \mapsto\ [0,1]
\end{equation}
telle que~:
\begin{eqnarray}
\sum_{A\subseteq \Omega} m^\Omega(A)=1
\end{eqnarray}
Un sous-ensemble de $2^\Omega$ de masse de croyance non-nulle est un élément focal. La masse affectée à un élément focal $A$ représente le degré de croyance élémentaire de la source à ce que la solution du problème soit $A$. Une fonction de masse permet ainsi de représenter des connaissances incertaines et imprécises d'une source d'informations. En général, nous manipulons des fonctions de masse non dogmatiques ({\em i.e.} dont l'ignorance $\Omega$ est élément focal), d'une part car elles permettent de systématiquement modéliser la part d'ignorance intrinsèque à toute source, mais également car toute fonction de masse non dogmatique est décomposable en fonctions de masse à support simple ({\em i.e.} qui ne comporte que deux éléments focaux dont $\Omega$). Les fonctions de masse à support simple sont notées $A^w$ telles que $m(A)=1-w$ $\forall A\neq\Omega$ et $m(\Omega)=w$. Ainsi une fonction de masse non dogmatique peut s'écrire~:
\begin{eqnarray}
m^\Omega=\displaystyle \ocap_{A\subset \Omega} A^{w(A)}
\end{eqnarray}
où $\ocap$ est données par l'équation~\eqref{conjunctive} ci-dessous.

La crédibilité $\bel$ et la plausibilité $\pl$ sont des fonctions duales définies à partir de la fonctions de masse et représentent respectivement une fonction de croyance minimale et maximale. Ainsi la fonction de plausibilité est donnée par~:
\begin{eqnarray}
\label{pl}
\pl(X)=\sum_{Y \subset \Omega, Y\cap X \neq \emptyset} m(Y)=\bel(\Omega)-\bel(X^c)=1-m(\emptyset)-\bel(X^c),
\end{eqnarray}
où $X^c$ est le complémentaire de $X$.

Une fois les fonctions de masse $m^\Omega_j$ déterminées pour chaque source d'informations $S_j$, plusieurs opérateurs de combinaison sont envisageables en fonction des hypothèses initiales. Les opérateurs de type conjonctif peuvent être employés lorsque les sources sont fiables et indépendantes cognitivement. La combinaison conjonctive s'écrit pour deux fonctions de masse $m^\Omega_1$ et $m^\Omega_2$ et pour tout $X \in 2^\Omega$ par~:
\begin{eqnarray}
\label{conjunctive}
m^\Omega_\conj(X)=m_1\ocap m_2= \displaystyle \sum_{Y_1\cap Y_2 =X} m^\Omega_1(Y_1)m^\Omega_2(Y_2).
\end{eqnarray}

Notons que l'élément neutre pour cette règle est la masse~: $m^\Omega_\Omega(X)=1$ si $X=\Omega$ et 0 sinon. Lorsque cette hypothèse de fiabilité est trop forte et que l'on ne peut supposer que seule une des sources est fiable, la combinaison disjonctive peut alors être employée toujours sous l'hypothèse d'indépendance cognitive~:
\begin{eqnarray}
\label{disjunctive}
m^\Omega_\dis(X)=\displaystyle \sum_{Y_1\cup Y_2 =X} m^\Omega_1(Y_1)m^\Omega_2(Y_2).
\end{eqnarray}

Notons que l'élément neutre pour cette règle est la masse~: $m^\Omega_\emptyset(X)=1$ si $X=\emptyset$ et 0 sinon. La plupart des règles de combinaison issues des règles conjonctives et disjonctives, en particulier pour répartir le conflit, supposent que les sources sont indépendantes cognitivement. \cite{Martin10a} en rappelle quelques unes. 

\cite{Denoeux08a} propose une famille de règles qui ne nécessitent pas l'hypothèse d'indépendance cognitive. Ainsi selon le comportement conjonctif ou disjonctif deux règles principales sont définies, la règle prudente et hardie. La règle prudente s'écrit pour les fonctions de masse non dogmatiques~:
\begin{eqnarray}
m^\Omega_1\owedge m^\Omega_2=\displaystyle \ocap_{A\subset \Omega} A^{w_1(A)\wedge w_2(A)}
\end{eqnarray}
où $\wedge$ est le maximum. La règle hardie s'écrit de même en considérant le minimum au lieu du maximum. Si ces règles sont efficaces lorsque les sources sont dépendantes, cette notion de dépendance ou d'indépendance n'est par clairement définie par \cite{Denoeux08a}.

Lorsqu'une connaissance supplémentaire garantie un sous-ensemble $A\subset \Omega$, nous pouvons définir une fonction de masse conditionnelle par~:
\begin{eqnarray}
m^\Omega[A](X)=(m^\Omega \ocap m^\Omega_A)(X)
\end{eqnarray}
où $m^\Omega_A(A)=1$ est la fonction de masse garantissant la réalisation de $A$.

\subsection{Notion d'indépendance}
\label{independance}
L'indépendance statistique est définie pour deux variables $A$ et $B$ par $P(A|B)=P(A)$ ou de façon équivalente par $P(A \cap B)=P(A) P(B)$. Cette indépendance est étendue par \cite{Shafer76} dans le cadre de la théorie des fonctions de croyance et est donnée par~: \linebreak $\pl(A \cap B)=\pl(A) \pl(B)$. \cite{BenYaghlane02a,BenYaghlane02b} définissent une indépendance doxatique entre des variables définies sur des cadres de discernement différents éventuellement. 

Ces définitions de l'indépendance ne correspondent pas à la notion d'indépendance cognitive entre les sources d'informations. Cette dernière se révèle très difficile à mesurer. \cite{Chebbah12a,Chebbah13a} proposent une définition d'une mesure d'indépendance entre deux sources d'informations étendues à une mesure de dépendance positive et négative. La mesure d'indépendance entre deux sources est définie comme une sorte de corrélation entre deux sources issues d'un {\em clustering} (classification non-supervisée) sur les fonctions de masse de chacune des sources en associant ensuite les clusters. Si $|\Omega|=n$, le clustering des fonctions de masse issues de la source $S_1$ fourni $n$ clusters, de même pour $S_2$. Les clusters des deux sources ($Cl_{k_1}$, $Cl_{k_2}$) sont associés de façon non symétrique en maximisant~:
\begin{equation}
\alpha^i_{k_1,k_2}=\frac{|Cl_{k_1}\cap Cl_{k_2}|}{|Cl_{k_i}|}, \, i=1,2
\end{equation}

Il est ensuite possible de définir une fonction de masse sur $\Omega_I=\{I,\bar{I}\}$ représentant les deux possibilités~: indépendant et dépendant ($\bar{I}$) de la source $S_1$ par rapport à la source $S_2$~:
\begin{equation}
\label{indCluster}
\left\{
\begin{tabular}{ll}
$m^{\Omega_I}_{{k_1}{k_2}}(I)=\beta \, (1-\alpha^1_{{k_1}{k_2}})$\\
$m^{\Omega_I}_{{k_1}{k_2}}(\bar{I})=\beta \, \alpha^1_{{k_1}{k_2}}$\\
$m^{\Omega_I}_{{k_1}{k_2}}(I \cup \bar{I})=1-\beta$\\
\end{tabular}
\right.
\end{equation}
où $\beta$ est un facteur d'affaiblissement permettant de tenir compte du nombre d'observations dans chaque cluster. Ainsi la croyance élementaire que la source $S_1$ est indépendante de $S_2$ est donnée par la masse~:
\begin{equation}
\label{ind}
m^{\Omega_I} (X)=\frac{1}{n}\left(\sum_{k_1=1}^n m^{\Omega_I}_{{k_1}{k_2}} \right)(X)
\end{equation} 
où $k_2$ est le cluster de la source $S_2$ associé au cluster $k_1$ de la source $S_1$. La moyenne est ici employée du fait de la dépendance des fonctions de masse.

\cite{Chebbah12a,Chebbah13a} proposent un prolongement pour différencier la dépendance positive (la source $S_1$ suit les avis de la source $S_2$) et la dépendance négative (la source $S_1$ dit le contraire des avis de la source $S_2$). Ainsi, une fonction de masse conditionnelle est construite sur le cadre de discernement $\Omega_P=\{P,\bar{P}\}$~:
\begin{equation}
\left\{
\begin{tabular}{ll}
$m^{\Omega_P}_{{k_1}{k_2}}[\bar{I}](P)=1-Dist(Cl_{k_1},Cl_{k_2})$\\
$m^{\Omega_P}_{{k_1}{k_2}}[\bar{I}](\bar{P})=Dist(Cl_{k_1},Cl_{k_2})$\\
$m^{\Omega_P}_{{k_1}{k_2}}[\bar{I}](P\cup \bar{P})=0$\\
\end{tabular}
\right.
\end{equation}
où $Dist(Cl_{k_1},Cl_{k_2})$ est la distance entre les deux clusters dépendants $Cl_{k_1}$ et $Cl_{k_2}$ liés comme étant la moyenne des distances entre les fonctions de masse des objets en commun~:
\begin{equation}
\label{eq11}
Dist(Cl_{k_1},Cl_{k_2})=\frac{1}{|Cl_{k_1}\cap Cl_{k_2}|}\sum_{j=1}^{|Cl_{k_1}\cap Cl_{k_2}|}d(m_{1,j}^{\Omega},m_{2,j}^{\Omega})
\end{equation}
où $d$ est la distance proposée par \cite{Jousselme01a} entre les fonctions de masse de la source $S_1$ et $S_2$ respectivement.

Si nous considérons que $\bar{I}=P\cup \bar{P}$, nous pouvons réécrire les deux fonctions de masse précédentes dans le cadre de discernement $\mathcal{I}=\{I,P,\bar{P}\}$. Nous définissons ainsi la fonction de masse entre deux clusters de $S_1$ et $S_2$~:
\begin{equation}
\label{m_indep}
\left\{
\begin{tabular}{ll}
$m^\mathcal{I}_{{k_1}{k_2}}(I)=m^{\Omega_I}_{{k_1}{k_2}}(I)=\beta \, \alpha^1_{{k_1}{k_2}}$\\
$m^\mathcal{I}_{{k_1}{k_2}}(P)=m^{\Omega_I}_{{k_1}{k_2}}(\bar{I}) m^{\Omega_P}_{{k_1}{k_2}}[\bar{I}](P)=\beta \, (1-\alpha^1_{{k_1}{k_2}})(1-Dist(Cl_{k_1},Cl_{k_2}))$\\
$m^\mathcal{I}_{{k_1}{k_2}}(\bar{P})=m^{\Omega_I}_{{k_1}{k_2}}(\bar{I}) m^{\Omega_P}_{{k_1}{k_2}}[\bar{I}](\bar{P})=\beta \, (1-\alpha^1_{{k_1}{k_2}}) Dist(Cl_{k_1},Cl_{k_2})$\\
$m^\mathcal{I}_{{k_1}{k_2}}(P\cup \bar{P})=m^{\Omega_I}_{{k_1}{k_2}}(\bar{I}) m^{\Omega_P}_{{k_1}{k_2}}[\bar{I}](P\cup \bar{P})=0 $\\
$m^\mathcal{I}_{{k_1}{k_2}}(I\cup P\cup \bar{P})=m^{\Omega_I}_{{k_1}{k_2}}(I \cup \bar{I})=1-\beta$\\
\end{tabular}
\right.
\end{equation}

La fonction de masse sur la dépendance de la source $S_1$ par rapport à $S_2$ est donnée par~:
\begin{equation}
\label{ind2}
m^\mathcal{I} (X)=\frac{1}{n}\left(\sum_{k_1=1}^n m^\mathcal{I}_{{k_1}{k_2}} \right)(X)
\end{equation} 
où $k_2$ est le cluster de la source $S_2$ associé au cluster $k_1$ de la source $S_1$. Cette fonction de masse représente ainsi l'ensemble des croyances élémentaires sur l'indépendance et dépendance positive et négative de la source $S_1$ face à la source $S_2$.

\subsection{Notion d'affaiblissement}
\label{affaiblissement}
 \cite{Shafer76} a proposé la procédure d'affaiblissement suivante~:
\begin{eqnarray}
^\alpha m^\Omega(X)&=&\alpha m^\Omega(X) \quad \forall X\in2^\Omega \setminus \Omega\\
^\alpha m^\Omega(\Omega)&=&1-\alpha (1-m^\Omega(\Omega))
\end{eqnarray}
où $\alpha$ est un facteur d'affaiblissement de $[0,1]$. Cette procédure est généralement employée pour affaiblir les fonctions de masse par la fiabilité $\alpha$ des sources d'informations. Cette procédure a pour effet d'augmenter la masse sur l'ignorance $\Omega$. \cite{Smets93a} a justifié cette procédure en considérant que~:
\begin{eqnarray}
m^\Omega[F](X)&=&m^\Omega(X)\\
m^\Omega[\bar{F}](X)&=&m^\Omega_\Omega(X)
\end{eqnarray}
où $m^\Omega_\Omega(X)=1$ si $X=\Omega$ et 0 sinon, $F$ et $\bar{F}$ représentent la fiabilité et la non fiabilité et $m^\Omega[F]$ est une fonction de masse conditionnelement à la fiabilité $F$. Soit $\mathcal{F} =\{F,\bar{F}\}$ le cadre de discernement correspondant, et la fonction de masse représentant la connaissance sur la fiabilité de la source~:
\begin{equation}
\label{fiab}
\left\{
\begin{tabular}{ll}
 $m^\mathcal{F}(F)=\alpha$\\
$m^\mathcal{F}(\mathcal{F})=1-\alpha$. \\ 
\end{tabular}
\right.
\end{equation}
Afin de combiner les deux sources d'informations fournissant les deux fonctions de masse $m^\Omega[F]$ et $m^\mathcal{F}$, il faut pouvoir les représenter dans le même espace $\Omega \times \mathcal{F}$. Ainsi, nous devons effectuer une {\em extention à vide} sur la fonction de masse $m^\mathcal{F}$, opération que l'on note $m^{\mathcal{F}\uparrow\Omega\times\mathcal{F}}$~:
\begin{equation}
m^{\mathcal{F}\uparrow\Omega\times\mathcal{F}} \left(Y\right)=\left\lbrace
\begin{array}{ll}
 m^\mathcal{F} \left(X\right)&\text{si } Y=\Omega \times X, \quad X\subseteq\mathcal{F}\\
0 &\text{sinon}
\end{array}
\right.
\end{equation}
Dans le cas de la fonction de masse $m^\Omega[F]$, il faut déconditionner~:
\begin{equation}
m^\Omega\left[F\right]^{\Uparrow\Omega\times\mathcal{F}}\left((A\times
F)\cup(\Omega\times\overline{F})\right)=m^\Omega\left[F\right]\left(A\right), \quad A\subseteq\Omega
\end{equation}
Il est ainsi possible d'effectuer la combinaison~:
\begin{equation}
m_\conj^{\Omega\times\mathcal{F}}(Y)= m^{\mathcal{F}\uparrow\Omega\times\mathcal{F}} \ocap m^\Omega\left[F\right]^{\Uparrow\Omega\times\mathcal{F}} (Y), \quad \forall Y\subset \Omega\times\mathcal{F}
\end{equation}
Ensuite il faut marginaliser la fonction de masse obtenue pour revenir dans l'espace $\Omega$~:
\begin{equation}
 m^{\Omega\times\mathcal{F}\downarrow\Omega}
\left(X\right)=\displaystyle{
\sum_{
\left\lbrace
Y\subseteq\Omega\times\mathcal{F} | Proj\left(Y\downarrow\Omega\right)=X
\right\rbrace}
m_\conj^{\Omega\times\mathcal{F}}\left(Y\right)
}
\end{equation}
où $Proj\left(Y\downarrow\Omega\right)$ est la projection de $Y$ sur $\Omega$. Nous retrouvons ainsi~:
\begin{equation}
 ^\alpha m^\Omega(X)=m^{\Omega\times\mathcal{F}\downarrow\Omega}\left(X\right)
\end{equation}

\cite{Mercier06a} a proposé une extension de cet affaiblissement en contextualisant le coefficient d'affaiblissement $\alpha$ en fonction de sous-ensembles de $\Omega$.

\section{Intégration de l'indépendance dans une fonction de masse}

Nous avons vu que la notion de l'indépendance est généralement une information supplémentaire nécessaire à la fusion d'informations, mais non prise en compte dans le formalisme choisi. La section~\ref{independance} propose une modélisation et estimation d'une mesure d'indépendance dans le cadre de la théorie des fonctions de croyance. Nous allons ici nous appuyer sur le principe de l'affaiblissement présenté dans la section~\ref{affaiblissement} afin de tenir compte de l'indépendance dans les fonctions de masse en vue de la combinaison. 

En effet, lors de la combinaison conjonctive par exemple l'hypothèse d'indépendance cognitive des sources d'informations est nécessaire. Si les sources ne sont pas indépendantes on peut penser qu'elles ne devraient pas être combinées par ce biais. Cependant, comme le montre la section~\ref{independance} les sources peuvent avoir des degrés de dépendance et d'indépendance. L'information fournie sur l'indépendance n'est pas catégorique. Ainsi, combiner deux sources indépendantes fortement devraient tendre vers le résultat de la combinaison de deux sources indépendantes. 
Si une source est dépendante d'une autre source, nous pouvons considérer que cette première source ne doit pas influer la combinaison avec une seconde source. Ainsi cette source doit représenter l'élément neutre de la combinaison. 

Dans ce cas, il suffit d'appliquer la procédure d'affiblissement de la section~\ref{affaiblissement} sur la fonction de masse $m^\Omega$ de la source $S_1$ en considérant l'indépendance donnée par la fonction de masse de l'équation~\eqref{ind} au lieu de celle de l'équation~\eqref{fiab} dans le cas de la fiabilité.

\`A présent, nous distinguons la dépendance positive de la dépendance négative. Si une source est dépendante positivement d'une autre source, il ne faut pas en tenir compte et donc tendre vers un résultat de combinaison qui prendrait cette première source comme un élément neutre. Enfin si une source est dépendante négativement d'une autre source, il peut être intéressant de marquer cette dépendance conflictuelle en augmentant la masse sur l'ensemble vide.

Pour réaliser ce schéma, nous proposons d'affaiblir les fonctions de masse d'une source $S_1$ en fonction de sa mesure d'indépendance à une autre source $S_2$, donnée par la fonction de masse $m^\mathcal{I}$ de l'équation~\eqref{ind2}. Nous réécrivons ici cette fonctions de masse de façon à simplifier les notations~:
\begin{equation}
\label{m_indep2}
\left\{
\begin{tabular}{ll}
$m^\mathcal{I}(I)= \alpha \beta$\\
$m^\mathcal{I}(P)=\alpha (1-\beta) \gamma$\\
$m^\mathcal{I}(\bar{P})=\alpha (1-\beta) (1-\gamma)$\\
$m^\mathcal{I}(I\cup P\cup \bar{P})=1-\alpha$\\
\end{tabular}
\right.
\end{equation}
Ainsi, le paramètre $\alpha$ représente la fiabilité de la source $S_1$, $\beta$ l'indédendance de $S_1$ face à $S_2$ et $\gamma$ la dépendance positive de $S_1$ face à $S_2$. Ces trois paramètres, $\alpha$, $\beta$ et $\gamma$ sont compris entre 0 et 1.

Nous considérons ici une fonction de masse d'une source $m^\Omega$ en fonction de son indépendance ou dépendance à une autre source. Ainsi nous définissons~:
\begin{equation}
\label{m_indep}
\left\{
\begin{tabular}{ll}
$m^\Omega[I](X)=m^\Omega(X)$\\
$m^\Omega[\bar{P}](X)=m^\Omega_\emptyset(X)$\\
$m^\Omega[P](X)=m^\Omega_\Omega(X)$\\
\end{tabular}
\right.
\end{equation}
où $m^\Omega_\Omega(X)=1$ si $X=\Omega$ et 0 sinon et $m^\Omega_\emptyset(X)=1$ si $X=\emptyset$ et 0 sinon. Suivant la procédure d'affaiblissement, nous effectuons une extension à vide sur la fonction de masse $m^\mathcal{I}$~:
\begin{equation}
m^{\mathcal{I}\uparrow\Omega\times\mathcal{I}} \left(Y\right)=\left\lbrace
\begin{array}{ll}
 m^\mathcal{I} \left(X\right)&\text{si } Y=\Omega \times X, \quad X\subseteq\mathcal{I}\\
0 &\text{sinon}
\end{array}
\right.
\end{equation}
Le déconditionnement des fonctions de masse $m^\Omega[I]$, $m^\Omega[P]$ et $m^\Omega[\bar{P}]$ est donné par~:
\begin{equation}
m^\Omega\left[I\right]^{\Uparrow\Omega\times\mathcal{I}}\left((A\times I)\cup(\Omega\times\overline{I})\right)=m^\Omega\left[I\right]\left(A\right), \quad A\subseteq\Omega
\end{equation}
où $\bar{I}=P\cup \bar{P}$.
\begin{equation}
m^\Omega\left[\bar{P}\right]^{\Uparrow\Omega\times\mathcal{I}}\left((A\times \bar{P})\cup(\Omega\times \{I\cup P\})\right)=m^\Omega\left[\bar{P}\right]\left(A\right), \quad A\subseteq\Omega
\end{equation}
\begin{equation}
m^\Omega\left[P\right]^{\Uparrow\Omega\times\mathcal{I}}\left((A\times P)\cup(\Omega\times \{I\cup\bar{P}\})\right)=m^\Omega\left[P\right]\left(A\right), \quad A\subseteq\Omega
\end{equation}
Ce dernier déconditionnement mène en fait à la masse de l'ignorance et est l'élément neutre de la combinaison conjonctive. 

Nous réalisons ensuite la combinaison conjonctive~:
\begin{equation}
m_\conj^{\Omega\times\mathcal{I}}(Y)= m^{\mathcal{I}\uparrow\Omega\times\mathcal{I}} \ocap m^\Omega\left[I\right]^{\Uparrow\Omega\times\mathcal{I}} \ocap m^\Omega\left[\bar{P}\right]^{\Uparrow\Omega\times\mathcal{I}} (Y), \quad \forall Y\subset \Omega\times\mathcal{I}
\end{equation}

La marginalisation de la fonction de masse permet ensuite de revenir dans l'espace $\Omega$~:
\begin{equation}
 m^{\Omega\times\mathcal{I}\downarrow\Omega}
\left(X\right)=\displaystyle{
\sum_{
\left\lbrace
Y\subseteq\Omega\times\mathcal{I} | Proj\left(Y\downarrow\Omega\right)=X
\right\rbrace}
m_\conj^{\Omega\times\mathcal{I}}\left(Y\right)
}
\end{equation}

Cette procédure réalisée pour la source $S_1$ en rapport à la source $S_2$ peut être réalisée pour la source $S_2$ au regard de la source $S_1$. Ainsi les deux fonctions de masse obtenue peuvent être combinées par la règle de combinaison conjonctive qui suppose l'indépendance. 

\section{Illustration}

\subsection{Fonctionnement de l'affaiblissement par la mesure d'indépendance}

Nous allons dans un premier temps illustrer le fonctionnement de l'affaiblissement par la mesure d'indépendance. Nous considérons ici un cadre de discernement $\Omega=\{\omega_1,\omega_2,\omega_3\}$. Supposons que nous ayons deux sources $S_1$ et $S_2$ donnant deux fonctions de masse~:
\begin{eqnarray}
\label{massesources1}
m_1^\Omega(\omega_1)=0.2, \, m_1^\Omega(\omega_1 \cup \omega_2)=0.5, \, m_1^\Omega(\Omega)=0.3,\\
\label{massesources2}
m_2^\Omega(\omega_2)=0.1, \, m_2^\Omega(\omega_1 \cup \omega_2)=0.6, \, m_2^\Omega(\Omega)=0.3
\end{eqnarray}
La combinaison conjonctive donne~:
\begin{eqnarray*}
&& m_{1\cap2}^\Omega(\emptyset)=0.02, \, m_{1\cap2}^\Omega(\omega_1)=0.18,\, m_{1\cap2}^\Omega(\omega_2)=0.08,\\ 
&& m_{1\cap2}^\Omega(\omega_1 \cup \omega_2)=0.63, \, m_{1\cap2}^\Omega(\Omega)=0.09
\end{eqnarray*}

Cette combinaison conjonctive est effectuée avec l'hypothèse d'indépendance cognitive des deux sources. Si une connaissance externe permet de mesurer la dépendance positive et négative de la source $S_1$ par rapport à la source $S_2$ telle que fournie par l'équation~\eqref{m_indep2}, nous devons en tenir compte avant la combinaison conjonctive. Supposons ainsi que $\alpha=0.95$, $\beta=0.05$ et $\gamma=0.95$ dans l'équation~\eqref{m_indep2}. Cette fonction de masse traduit donc une forte dépendance positive de $S_1$ par rapport à $S_2$. Nous avons ainsi la fonction de masse~:
\begin{equation}
\label{m_indep2}
\left\{
\begin{tabular}{ll}
$m^\mathcal{I}(I)= 0.0475$\\
$m^\mathcal{I}(P)=0.8574$\\
$m^\mathcal{I}(\bar{P})=0.0451$\\
$m^\mathcal{I}(I\cup P\cup \bar{P})=0.05$\\
\end{tabular}
\right.
\end{equation}

Le tableau~\ref{details} présente les différentes étapes d'extension à vide, de déconditionnement et de combinaison dans l'espace $\Omega\times\mathcal{I}$. L'extension à vide et le déconditionnement transfèrent les masse sur les éléments focaux correspondant de l'espace $\Omega\times\mathcal{I}$. La combinaison des trois fonctions de masse dans cet espace fait apparaître la masse sur l'ensemble vide qui correspond à la part de dépendance négative.
 
\begin{table}[h!]
\begin{center}
\begin{tabular}{|c|c|c|c|c|}
\hline
 & & & & \\
focal & $m^{I\uparrow\Omega\times\mathcal{I}}$ & $m^\Omega[I]^{\Uparrow\Omega\times\mathcal{I}}$ & \!\!$m^\Omega[\bar{P}]^{\Uparrow\Omega\times\mathcal{I}}$\!\! & $m_\conj^{\Omega\times\mathcal{I}}$\\
\hline
\vspace{-0.3cm} &  & & &\\
$\emptyset$ & & &  & 0.0451\\
\hline
\vspace{-0.3cm} &  & & &\\
$\omega_1 \times I$ &  & & & 0.0095\\
\hline
\vspace{-0.3cm} &  & & &\\
$(\omega_1 \cup \omega_2) \times I$ &  & & & 0.0237\\
\hline
\vspace{-0.3cm} &  & & &\\
$\Omega \times I$ & 0.0475 & & & 0.0142\\
\hline
\vspace{-0.3cm} &  & & &\\
$\Omega \times P$ & 0.8574 & & & 0.8574\\
\hline
\vspace{-0.3cm} &  & & &\\
$(\omega_1 \times I) \cup (\Omega \times P) $ &  & & &0.01\\
\hline
\vspace{-0.3cm} &  & & &\\
$((\omega_1 \cup \omega_2) \times I) \cup (\Omega \times P)$ &  & & & 0.025\\
\hline
\vspace{-0.3cm} &  & & &\\
$\Omega \times \bar{P}$ & 0.0451 & & &\\
\hline
\vspace{-0.3cm} &  & & &\\
$\Omega \times (P \cup  \bar{P})$ &  & & 1&\\
\hline
\vspace{-0.3cm} &  & & &\\
$(\omega_1 \times I) \cup (\Omega \times (P \cup  \bar{P}))$ &  & 0.2& &\\
\hline
\vspace{-0.3cm} &  & & &\\
$((\omega_1 \cup \omega_2) \times I) \cup (\Omega \times (P \cup  \bar{P}))$ &  &0.5 & &\\
\hline
\vspace{-0.3cm} &  & & &\\
$\Omega \times \mathcal{I}$ & 0.05 & 0.3 & & \\
\hline 
\end{tabular}
\caption{Détails de l'affaiblissement de la mesure d'indépendance~: fonctions de masse dans $\Omega\times\mathcal{I}$.}
\label{details}

\end{center}
\end{table}

Le tableau~\ref{details2} présente ensuite la marginalisation et le résultat de combinaison avec la fonction de masse $m_2$ non modifiée ({\em i.e.} que l'hypothèse d'indépendance totale de $S_2$ par rapport à $S_1$ est faite). Nous constatons que la masse transférée sur l'ignorance devient plus importante que lors de la combinaison conjonctive sans hypothèse sur la dépendance positive.

\begin{table}[h!]
\begin{center}
\begin{tabular}{|c|c|c|c|}
\hline
focal & $m_1^{\Omega\times\mathcal{I} \downarrow \Omega}$ & $m_2^{\Omega}$ & $m_1^{\Omega\times\mathcal{I} \downarrow \Omega} \ocap m_2^{\Omega}$\\
\hline
$\emptyset$ & 0.0451 & & 0.0461\\
\hline
$\omega_1$ & 0.0095 &  & 0.0085\\
\hline
$\omega_2$ & & 0.1 & 0.0945\\
\hline
$\omega_1 \cup \omega_2$ & 0.0237 & 0.6 & 0.5743\\
\hline
$\Omega$ & 0.9216 & 0.3 & 0.2765 \\
\hline 
\end{tabular}
\caption{Détails de l'affaiblissement de la mesure d'indépendance~: marginalisation et combinaison}
\label{details2}

\end{center}
 
\end{table}

\begin{figure}[h!]
\begin{center}
\includegraphics[height=5cm]{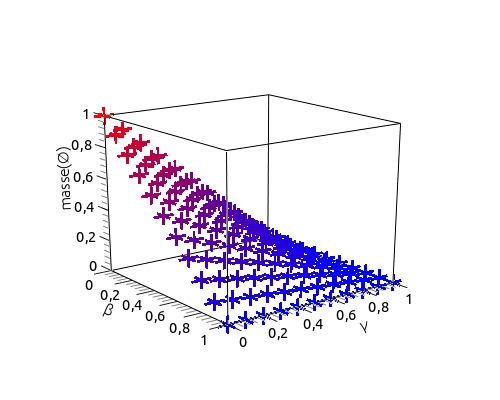}
\end{center}
\caption{Variation de la masse sur l'ensemble vide pour un affaiblissement par la mesure d'indépendance d'une masse dogmatique.}
\label{fig1}
\end{figure}

\begin{figure}[h!]
\begin{center}
\includegraphics[height=5cm]{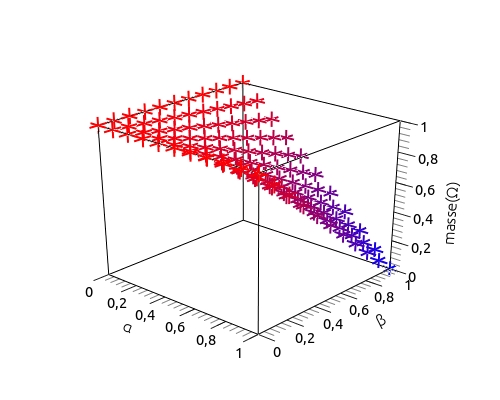}
\end{center}
\caption{Variation de la masse sur l'ignorance $\Omega$ pour un affaiblissement par la mesure d'indépendance d'une masse dogmatique.}
\label{fig2}
\end{figure}

Afin de bien illustrer le transfert de masse sur l'ensemble vide et sur l'ignorance, les figures~\ref{fig1} et \ref{fig2} représentent les masses en fonction des variations de $\alpha$, $\beta$ et $\gamma$ pour une fonction de masse dogmatique quelconque. Ainsi sur la figure~\ref{fig1} représentant les variations de masse sur l'ensemble vide, $\alpha$ est fixé à 1, $\beta$ et $\gamma$ variant, alors que sur la figure~\ref{fig2} représentant les variations de masse sur l'ignorance, $\gamma$ est fixé à 1, $\alpha$ et $\beta$ variant.

La figure~\ref{fig1} montre ainsi que plus $\beta$ et $\gamma$ sont petits plus on obtient une masse importante sur l'ensemble vide et donc une dépendance négative. La quantité $\beta$ représente la part d'indépendance et la quantité $\gamma$ représente la part de dépendance positive.

La figure~\ref{fig2} présente quand à elle, la variation de la masse sur $\Omega$, l'ignorance. Cette masse est donnée directement par $\alpha \beta$ qui contient donc la part d'indépendance $\beta$ et la fiabilité $\alpha$ de la source. 

Nous illustrons ainsi le résultat escompté de l'affaiblissement par la mesure d'indépendance, c'est-à-dire que nous retrouvons sur la masse de l'ensemble vide la quantité de dépendance négative et sur l'ignorance la quantité de fiabilité et de d'indépendance.

\subsection{Influence sur le résultat de combinaison}
Afin d'illustrer l'influence de la prise en compte de la mesure d'indépendance sur les fonctions de masse, nous allons considérer ici les deux sources précédentes $S_1$ et $S_2$ qui fournissent les fonctions de masse données par les équations~\eqref{massesources1} et \eqref{massesources2}. Nous allons considérer trois cas pour chaque source avec un cas où la source $S_1$ est plutôt indépendante de $S_2$ ($\alpha=0.95$, $\beta=0.95$, $\gamma=0.05$), un cas où elle est plutôt dépendante positivement ($\alpha=0.95$, $\beta=0.05$, $\gamma=0.95$) et un cas où elle est plutôt dépendante négativement ($\alpha=0.95$, $\beta=0.05$, $\gamma=0.05$). Pour la source $S_2$ nous considérons trois cas moins catégorique en fixant la fiabilité $\alpha=0.9$~: le cas plutôt indépendant ($\beta=0.9$, $\gamma=0.1$), le cas plutôt dépendant positivement ($\beta=0.1$, $\gamma=0.9$) et le cas plutôt dépendant négativement ($\beta=0.1$, $\gamma=0.1$).

\begin{table}[ht]
\begin{center}
\begin{tabular}{|c|c|c|c|c|c|c|c|c|}
\hline
 & & & \multicolumn{6}{c|} {$S_2$ : $\alpha=0.9$} \\
\cline{4-9}
cas & élément &  & \multicolumn{2}{c|}{$\beta=0.9$, $\gamma=0.1$} & \multicolumn{2}{c|}{$\beta=0.1$, $\gamma=0.9$} & \multicolumn{2}{c|}{$\beta=0.1$, $\gamma=0.1$}\\

\cline{4-9}
\vspace{-0.3cm} &  & & & & & & & \\
 & focal & \!\!\!\! $m_1^{\Omega\times\mathcal{I}\downarrow\Omega}$\!\!\!\! &\!\!\!\!  $m_2^{\Omega\times\mathcal{I}\downarrow\Omega}$ \!\!\!\!&  \!\!\!\!$m_{1\cap2}^{\Omega\times\mathcal{I}\downarrow\Omega}$\!\!\!\! & \!\!\!\!  $m_2^{\Omega\times\mathcal{I}\downarrow\Omega}$ \!\!\!\!& \!\!\!\! $m_{1\cap2}^{\Omega\times\mathcal{I}\downarrow\Omega}$\!\!\!\! & \!\!\!\! $m_2^{\Omega\times\mathcal{I}\downarrow\Omega}$\!\!\!\! &\!\!\!\!  $m_{1\cap2}^{\Omega\times\mathcal{I}\downarrow\Omega}$\!\!\!\! \\
\hline
$S_1$ & $\emptyset$ & 0.0451 & 0.081 & 0.1371 &  0.081 & 0.1240 &0.729 &0.7428 \\
\cline{2-9}
$\alpha=0.95$ & $\omega_1$ & 0.1805 & & 0.1513 & & 0.1643 & &0.0473 \\
\cline{2-9}
 $\beta=0.95$ & $\omega_2$ &  &0.081 & 0.0627 & 0.009 & 0.007 &0.009 &0.007 \\
\cline{2-9}
 $\gamma=0.05$ & $\omega_1 \cup \omega_2$ & 0.4513  &0.486 & 0.5352 &0.054 & 0.4281 &0.054 & 0.1357 \\
\cline{2-9}
 & $\Omega$ & 0.3231 & 0.352 & 0.1137 & 0.856 & 0.2766 & 0.208 & 0.0672 \\
\hline
\hline
$S_1$ & $\emptyset$ & 0.0451 & 0.081 & 0.1232 & 0.081 & 0.1226 &0.729 & 0.7413 \\
\cline{2-9}
$\alpha=0.95$ & $\omega_1$ & 0.0095 & & 0.008 & & 0.0086 & & 0.0025 \\
\cline{2-9}
 $\beta=0.05$ & $\omega_2$ &  &0.081 & 0.0766 & 0.009 &0.0085 & 0.009 & 0.0085 \\
\cline{2-9}
 $\gamma=0.95$ & $\omega_1 \cup \omega_2$ & 0.0238  &0.486 & 0.4678 &0.0054 & 0.0714 & 0.054 & 0.056 \\
\cline{2-9}
 & $\Omega$ & 0.9216 & 0.352 & 0.3244 & 0.856 & 0.7889 & 0.208 &0.1917 \\
\hline
\hline
$S_1$ & $\emptyset$ & 0.8574 & 0.081& 0.8697 & 0.081& 0.869 &0.729 &0.9614 \\
\cline{2-9}
$\alpha=0.95$ & $\omega_1$ & 0.0095 & & 0.008 & & 0.0087 &  & 0.0025 \\
\cline{2-9}
 $\beta=0.05$ & $\omega_2$ &  &0.081 & 0.0108 & 0.009& 0.0012 &0.009 & 0.0012 \\
\cline{2-9}
 $\gamma=0.05$ & $\omega_1 \cup \omega_2$ & 0.0237  &0.486 & 0.073 & 0.054 & 0.0275 & 0.054 & 0.0121\\
\cline{2-9}
 & $\Omega$ & 0.1094 & 0.352 & 0.0385 &0.856 & 0.0936 & 0.208 & 0.0228 \\
\hline
\end{tabular}
\caption{Résultats de combinaison selon les hjypothèses de dépendance et d'indépendance des deux sources $S_1$ et $S_2$.}
\label{res_comb}

\end{center}
\end{table}

Ainsi, le tableau~\ref{res_comb} présente les résultats de la combinaison des deux sources en fonction des hypothèses d'indépendance et de dépendance, positive ou négative des deux sources $S_1$ et $S_2$. Nous constatons que lorsque les deux sources sont plutôt indépendantes l'une de l'autre, les résultats obtenus sont proches de ceux obtenus par la combinaison conjonctive directe sous l'hypothèse d'indépendance. Lorsqu'une des deux sources est dépendante négativement de l'autre, la masse transférée sur l'ensemble vide est importante. Lorsque l'une des deux sources est dépendante positivement la masse transférée sur l'ignorance mais de façon moins importante que pour la dépendance négative. En effet, l'ensemble vide est un élément absorbant pour la combinaison conjonctive. Cette masse sur l'ensemble vide à l'issue de la combinaison conjonctive peut ainsi jouer un rôle d'alerte sur la dépendance négative. Une autre alternative serait d'envisager une autre règle de combinaison lorsque la masse issue de la 
dépendance négative est trop importante.

\section{Conclusion}

Cet article souligne l'importance de mesurer la réelle indépendance des sources et d'en tenir compte en vue de la combinaison des informations issues de celles-ci. Nous nous restreignons ici à la théorie des fonctions de croyance qui représente un cadre assez général pour la fusion d'informations. De ce contexte théorique, nous avons montré une approche originale pour intégrer une mesure d'indépendance exprimée sous la forme d'une fonction de masse. Nous avons ensuite explicité les fonctions de masse conditionnellement à leur indépendance mutuelle par un procédé d'affaiblissement des masses initiales. Ce principe doit être réalisé en vue d'une combinaison de ces fonctions de masse qui nécessite l'hypothèse d'indépendance. 

Une autre approche envisageable serait d'intégrer la mesure d'indépendance dans la combinaison des fonctions de masse.

\bibliographystyle{rnti}
\bibliography{biblio}
\end{document}